\documentclass{article}
\usepackage{spconf,amsmath,graphicx}

\def\etal{\emph{et al}. }
\def\eg{\emph{e.g}.}

\title{DEEP MULTI-SCALE ARCHITECTURES FOR MONOCULAR DEPTH ESTIMATION}

\name{M. Moukari$^{1,2}$, S. Picard$^{1}$, L. Simon$^{2}$, F. Jurie$^{2}$}
\address{(1) Safran -- (2) Normandie Univ, UNICAEN, ENSICAEN, CNRS (UMR GREYC)}
\begin{document}

\maketitle

\begin{abstract}

This paper aims at understanding the role of  multi-scale information in the estimation of depth from monocular images. More precisely, the paper investigates four different deep CNN architectures, designed to explicitly make use of multi-scale features along the network, and compare them to a state-of-the-art single-scale approach. The paper also shows that involving multi-scale features in depth estimation not only improves the performance in terms of accuracy, but also gives qualitatively better depth maps. Experiments are done on the widely used NYU Depth dataset, on which the proposed method achieves state-of-the-art performance.

\end{abstract}


\section{Introduction}
\label{sec:intro}

The estimation of depth information from images has a very long history in the computer vision literature, for instance in stereo vision \cite{scharstein2001}. As mentioned by Michels \etal \cite{michels2005high} many researchers have been inspired by the way humans use monocular cues (\eg texture, perspective, defocus) for estimating depth information \eg, \cite{Wu2004,Bulthoff1998}.

One major breakthrough in this area arose from the use of Deep Convolutional Networks, in order to train filters capable of detecting optimal cues for depth estimation. Mohan \cite{mohan2014deep} and Eigen \etal \cite{eigen2014depth} were among the firsts to propose depth estimation algorithms using deep CNN. Such algorithms now constitute the mainstream approach for depth estimation.  
Numerous publications have extended this approach in several directions. However, none of them scrutinize the importance of multi-scale information, in particular regarding the performance of  depth estimation.  This paper aims at filling this gap by proposing and evaluating several architectures that are multi-scale by construction and evaluate them on a public dataset. Our experimentations allowed us to observe that if multi-scale architectures bring  only mild improvements over careful designed single-scale architectures with small training sets, they lead to state-of-the-art results on larger ones.

The rest of the paper is structured as follows. Section \ref{sec:related}  presents the related works, Section~\ref{sec:method} the different  multi-scale architectures while Section \ref{sec:experiments} experimentally compares them and draw some conclusions.

\begin{figure}
\centering
\includegraphics[scale=0.6]{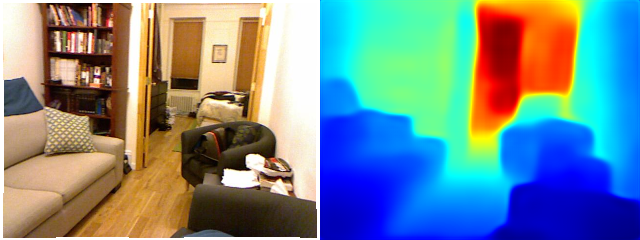}
\vspace{-1em}\caption{An example of our state-of-the-art monocular depth estimation trained with a deep multi-scale architecture.}\vspace{-1em}
\label{estimations}
\end{figure}


\section{Related Work}
\label{sec:related}
A number of deep learning methods have recently emerged to address the problem of monocular depth estimation. The seminal work of Eigen \etal \cite{eigen2014depth} introduced a powerful deep CNN approach allowing to estimate depth  from  monocular RGB images in a feed-forward way.  The proposed architecture is composed of two stacked networks. The first one outputs a coarse depth map while the second one refines it by applying 3 successive convolutions, using the coarse output and the RGB image as inputs.

This work has been extended in \cite{eigen2015predicting} by adding a third scale and by combining several tasks (multi-task learning framework), predicting not only the depth but also the surface normals and the semantic labels of the input RGB images. Several authors (\eg, \cite{wang2015towards}) have also tried to combine complementary tasks, such as semantic segmentation, to improve the estimation of the depth. 

In practice, recent methods can be roughly categorized into two main families: global methods that estimate the depth map from the whole input image and local methods that work locally on image patches. For instance, Eigen \etal \cite{eigen2014depth,eigen2015predicting} are both global methods. On the other hand, approaches such as \cite{liu2016learning, liu2015deep, li2015depth, roy2016monocular} are fundamentally based on local predictions. Some of them \cite{liu2016learning, liu2015deep, li2015depth} enforce some global consistency between patches thanks to CRF, yet they still remain fundamentally local. Furthermore, to alleviate the cons of both categories, Wang \etal \cite{wang2015towards} and Chakrabarti \etal \cite{chakrabarti2016depth} have designed hybrid approaches that fuse local and global predictions.

As we write, the state-of-the-art architecture for depth estimation is an encoder-decoder network with a powerful encoder typically pre-trained for semantic purposes. For example, Laina \etal\cite{laina2016deeper} use the ResNet architecture as a feature extractor, after cutting off the last classification layers and training a decoder on top of it to achieve the regression task of depth prediction. They propose up-sampling layers based on the principle of ResNet with skip connections and also show that training with the BerHu loss can further improve the final error. They achieve state-of-the-art performance on depth estimation from single monocular images. Although their work yields impressive results, they do not integrate multi-scale analysis along the network to enhance their prediction.

Convolutional architectures that are designed to use multi-scale features all along the network are now well studied in some research areas such as object detection, object recognition and semantic segmentation. 
 The U-net \cite{ronneberger2015u} is a popular encoder-decoder architecture where skip connections allow the decoder to benefit from the mirror encoder features, to help awareness of the finer details. Furthermore, Lin \etal \cite{lin2016feature} argue that, in addition to skip connections, it is useful to have some intermediate outputs at lower resolution in the decoder, constructing high-level semantic feature maps at different scales. They demonstrate that, at last, it allows to improve both object detection and segmentation tasks. Eventually, dilated convolutions have been introduced to avoid down sampling in the encoding part of the network while preserving the same receptive field. These layers have been exploited for semantic segmentation in a recent work \cite{chen2017rethinking} to implement deep feature extraction in a so-called spatial pyramid.

\section{Multi-scale architectures for depth estimation}
\label{sec:method}

\begin{figure}
\centering
\includegraphics[scale=0.31]{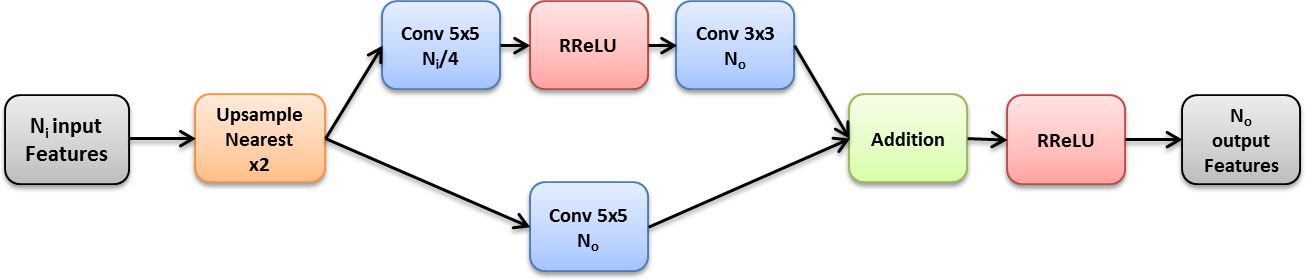}
\vspace{-1em}\caption{Our modified up-projection module. Unlike the original proposed in \cite{laina2016deeper}, we perform nearest neighbors up-sampling x2 and we use randomized ReLU.}\vspace{-1em}
\label{upprojection}
\end{figure}

\begin{figure}
\centering
\includegraphics[scale=0.27]{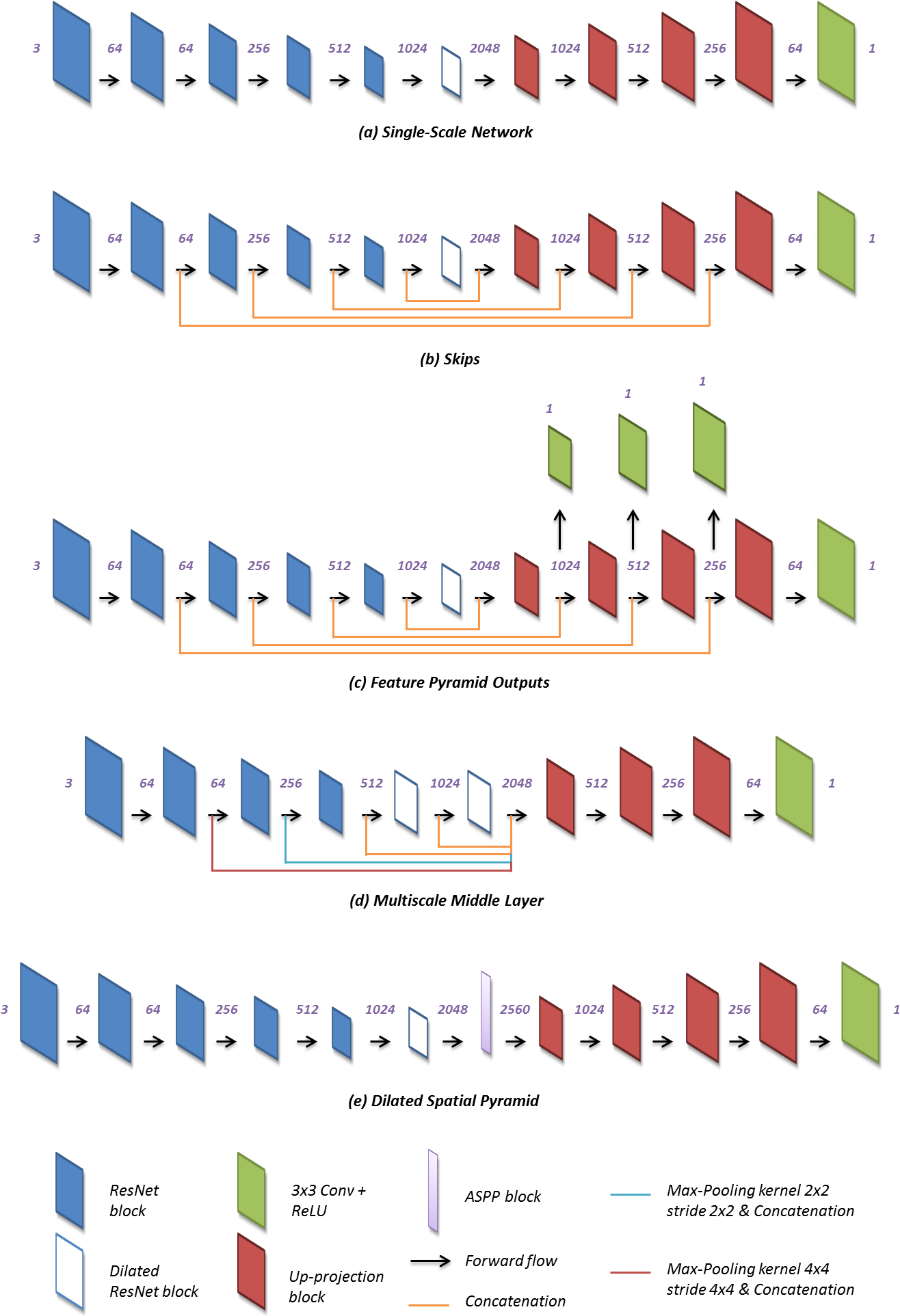}
\vspace{-1em}\caption{Schematics of the different compared methods.}\vspace{-1em}
\label{architectures}
\end{figure}

Given the lack of literature on the proper handling of multi-scale features for depth estimation, we propose here four architectures inspired from other research areas such as semantic segmentation. All four of them are built upon a highly optimized single-scale network that we present first. The five resulting network architectures are depicted in Fig.~\ref{architectures}).

\vspace{1em}\noindent \textbf{\bf Single-scale network (SSN)} This architecture is similar to hourglass nets, combining an encoder and a decoder part. We use the ResNet200 network \cite{he2016deep}, pre-trained on the ImageNet classification task, as a feature extractor. The last two layers, originally dedicated to classification purposes, are removed. Furthermore, we modify the vanilla convolutions of the last ResNet block to become 2-dilated convolutions. In this case, the output features have a higher spatial resolution, while maintaining the same receptive field. This type of strategy has been denoted as the {\em Dilated ResNet}  by  Yu \etal \cite{yu2017dilated}. Thereby, our dilated ResNet200 encoder outputs 2048 feature maps at 1/16 of the size of the input image.\\
For the decoding part, we use a slightly different version of the up-projection module, originally proposed in the work of Laina \etal \cite{laina2016deeper}. In our version (Fig.~\ref{upprojection}), we choose to use the nearest neighbors up-sampling instead of the original unpooling layer because we found it faster, with no impact on the final performance. Furthermore, as we conduct experiments on few training images, we use randomized ReLUs as activation units instead of the original ReLUs. We found it to generalize better when we learn on small-scale training set (typically less than 800 images). Similar results are observed in \cite{xu2015empirical}, reporting the effectiveness of randomized ReLUs to combat overfitting on small-scale datasets. At the far end of the network, we use a 3x3 convolution with 1 output channel, followed by a ReLU. The output of this last layer serves as our depth estimate.

\vspace{1em}\noindent \textbf{\bf Skip Connections (Skips)}  This first multi-scale variant is inspired by the U-Nets \cite{ronneberger2015u}: we use skip connections between mirror layers (in terms of spatial resolution) to concatenate features from the encoder with their counterpart of the same size in the decoder (see Fig. \ref{architectures} (b)).

\vspace{1em}\noindent \textbf{\bf Feature Pyramid Outputs (FPO)} This architecture is inspired by the work of Lin \etal \cite{lin2016feature}. It is the same architecture as the one used for \textit{Skips} but with additional extra intermediate outputs after the up-projection modules. These extra outputs are designed as a fork of the network and each contains one 3x3 convolution followed by a ReLU, allowing depth estimation at lower scales (see Fig. \ref{architectures} (c)).
We experimentally find more efficient to train each output sequentially, starting with the lower resolution and fixing the previously trained part of the network to learn a new output resolution.

\vspace{1em}\noindent \textbf{\bf Multi-Scale Middle Layer (MSML)}  This architecture also uses an encoder based on the Dilated ResNet. However, the last two ResNet blocks are replaced by 2 and 4 dilated ResNet blocks. Thus, we keep a higher spatial resolution in the middle layer whose size becomes 1/8 of the input resolution. Then, we gather all same-sized multi-scale encoder features in the middle layer before building on the decoder. Early features that are not the same size are first forwarded through a max pooling layer to downsample them before concatenation (see Fig. \ref{architectures} (d)).

\begin{figure}
\centering
\includegraphics[scale=0.36]{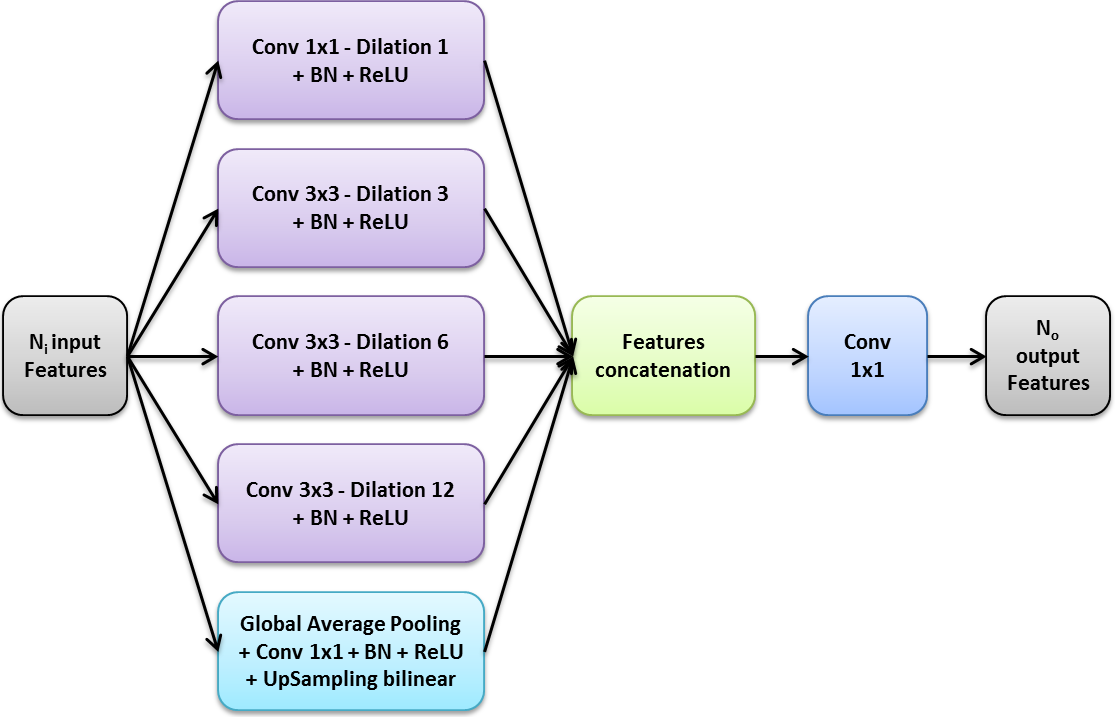}
\vspace{-1em}\caption{ASPP block (adapted from \cite{chen2017rethinking}), with a pyramid of 1, 3, 6 and 12-dilated convolutions. BN stands for Batch Normalization.}\vspace{-1em}
\label{aspp}
\end{figure}

\vspace{1em}\noindent \textbf{\bf Dilated Spatial Pyramid (DSP)} It relies on the same Atrous Spatial Pyramid Pooling (ASPP) blocks as the ones introduced by Chen \etal \cite{chen2017rethinking} (see Fig. \ref{architectures} (e)). The decoder part is build on top of the ASPP block \cite{chen2017rethinking}. We use a pyramid of 1, 3, 6 and 12-dilated convolutions to construct our ASPP block instead of the originals 1, 6, 12 and 18-dilations to adapt to the size of our features, as illustrated in Fig. \ref{aspp}.

\vspace{1em}\noindent \textbf{Decoder designs}
Between the five architectures, the decoder part varies slightly in the number of up-projections involved to recover the same spatial size as the input, and also in the amount of input and output features in each module. Hence, we use the following syntax to shorten the notations: [$N_i$-$N_o$]. Each bracket represents one up-projection module with $N_i$ input features and $N_o$ output features. All decoders up-projections architectures are summarized in Table \ref{decoders}.

\begin{table}
\begin{center}
\resizebox{\columnwidth}{!}{%
\begin{tabular}{|c||c|c|c|c|}
\hline
\multicolumn{5}{|c|}{\textbf{Decoders up-projections (UpProj) architectures}}\\
\hline\hline
Method & UpProj1 & UpProj2 & UpProj3 & UpProj4\\
\hline\hline

SSN & [2048-1024] & [1024-512] & [512-256] & [256-64]\\\hline
Skips& [3072-1024] & [1536-512] & [768-256] & [320-64]\\
FPO & [3072-1024] & [1536-512] & [768-256] & [320-64]\\
MSML & [3904-512] & [512-256] & [256-64] & -\\
DSP & [2560-1024] & [1024-512] & [512-256] & [256-64]\\

\hline
\end{tabular}%
}
\end{center}
\vspace{-1.5em}\caption{Architecture of the different up-projections modules for each experiment.}\vspace{-1em}
\label{decoders}
\end{table}
\section{Experiments}
\label{sec:experiments}

This section presents an experimental validation of the different architectures proposed in the previous section. These models are evaluated both quantitatively and qualitatively and compared with the recent state-of-the-art results. 

\vspace{1em}\noindent {\bf The NYU Depth v2 dataset.} All our experiments have been conducted on the popular and challenging NYU Depth benchmark~\cite{Silberman:ECCV12}. This dataset is composed of 464 video sequences of RGB-D indoor images, acquired with a Microsoft Kinect camera. The spatial resolution is of 640x480 pixels. The depth range goes from 0 to 10 meters. The dataset is split into 249 training sequences and 215 test sequences, among which only 795 training images and 654 testing images are densely annotated with filled-in depth values \cite{Silberman:ECCV12}. For both training and testing, we down sample the images by a factor of 2, using the nearest-neighbor algorithm.

\vspace{1em}\noindent {\bf Extended NYU dataset.} In the prospect of comparing our approaches with state-of-the-art result, we have extended the official NYU dataset. To do so, we randomly select equally spaced frames from the raw training scene of the NYU Depth dataset. We end up with about 16K unique images that we process using the depth colorization routine provided in the matlab NYU Depth toolbox. Similar unofficial training sets are exploited in recent published work such as \cite{laina2016deeper}.

\begin{table}
\begin{center}
\begin{tabular}{|c||c|c|c||c|c|c|}
\hline
& \multicolumn{3}{|c||}{\textbf{lower is better}} & \multicolumn{3}{|c|}{\textbf{higher is better}}\\
\hline\hline
Method & rel & log & rms & $\delta_1$ & $\delta_2$ & $\delta_3$ \\
\hline\hline

SSN & 0.167 & 0.066 & 0.603 & 77.9 & 94.9 & 98.9 \\\hline
Skips & 0.163 & 0.065 & 0.599 & 79.2 & 95.1 & 98.9 \\
MSML & 0.160 & 0.066 & 0.601 & 78.1 & \textbf{95.4} & \textbf{99.0} \\
FPO & 0.161 & 0.065 & 0.599 & 78.5 & 95.3 & 98.9 \\
DSP & \textbf{0.159} & \textbf{0.064} & \textbf{0.592} & \textbf{79.3} & \textbf{95.4} & 98.9 \\

\hline
\end{tabular}
\end{center}
\vspace{-1.5em}\caption{Error comparison of the multi-scale methods on the NYU Depth v2 test set. We only use the official subset of 795 training images for the learning stage.}\vspace{-1em}
\label{our_methods}
\end{table}

\vspace{1em}\noindent {\bf Details on the learning procedure} 
To obtain as fair comparisons as possible, we train our different architectures with the same hyper parameters and training procedures. We always rely on L2 losses and Stochastic Gradient Descent with a batch of size 3 (except for MSML and the last 2 scales of FPO where, due to memory limitations, we have used batches of size 2). We train each network for at least 800 epochs and stop the training when the score on the validation set increases during 50 epochs.
The learning rate is set to 5e-3, the weight decay to 5e-4 and momentum to 0.9. We also apply online data augmentation, following the procedure reported in \cite{eigen2014depth}.

\vspace{1em}\noindent {\bf Evaluation criteria.} We report the following four standard performance criteria ($\hat{y_i}$ and $y_i$ denote respectively the predicted and ground-truth depth at pixel $i$. $N$ stands for the total number of pixels):\\
(a) Mean relative error (rel): $\frac{1}{N}\sum_{i=1}^{N} \left |  \hat{y_{i}} - y_{i} \right | / y_{i}$\\
(b)  Mean $log_{10}$ error (log): $\frac{1}{N}\sum_{i=1}^{N}\left | log_{10} \ \hat{y_{i}}-log_{10} \ y_{i} \right |$\\
(c) RMSE (rms): $\sqrt{\frac{1}{N}\sum_{i=1}^{N} \left ( \hat{y_{i}} - y_{i} \right )^{2}}$\\
(d) Threshold: \% of $y_i$ s.t. $max \left ( \frac{\hat{y_{i}}}{y_{i}}, \frac{y_{i}}{\hat{y_{i}}} \right ) = \delta < thr$. In our evaluations, $\forall k\in\{1,2,3\}$, the $\delta_k$ stands for the threshold metric $\delta<1.25^k$.

\vspace{1em}\noindent {\bf Experiments on the Official NYU dataset.}
As a first analysis stage, we trained all our architectures on the small official NYU training set (795 images). The results are presented in Table~\ref{our_methods}.
In regard to the multi-scale vs single-scale comparison, we notice that every multi-scale network performs similarly or better than SSN. Though the performance gap is not so pronounced, we can conclude that for such a small dataset a carefully designed architecture as SSN is quite competitive.
Considering now the multi-scale methods, DSP consistently outperforms the other challengers. As a result, DSP stands out as a promising design to achieve state-of-the-art results.

\vspace{1em}\noindent {\bf Experiment on the Extended NYU dataset.}
Given our previous analysis, we singled out the DSP method for training on the unofficial but much larger scale NYU dataset. The learning procedure is the same, except that we train for about 60 epochs. The results are reported in Table~\ref{sota_methods}, along with recent state-of-the-art performance. We see that our proposed Large Scale DSP (LS-DSP) network achieves the best results for almost all the usual metrics.
Please note that, in the table, the results of Laina \etal~\cite{laina2016deeper} were deliberately restricted to the L2 loss while the authors recommend the berHu loss. This intentional choice was made to mitigate the effect of the loss in our comparison. Nonetheless, for the sake of fairness, we should mention that Laina \etal\cite{laina2016deeper} obtain slightly better results on \emph{rel} and \emph{log} errors, while we achieve better results on the 4 remaining metrics. We expect that in the case of DSP as well, the berHu loss could further improve the performance. 

\begin{figure}
\centering
\includegraphics[scale=0.26]{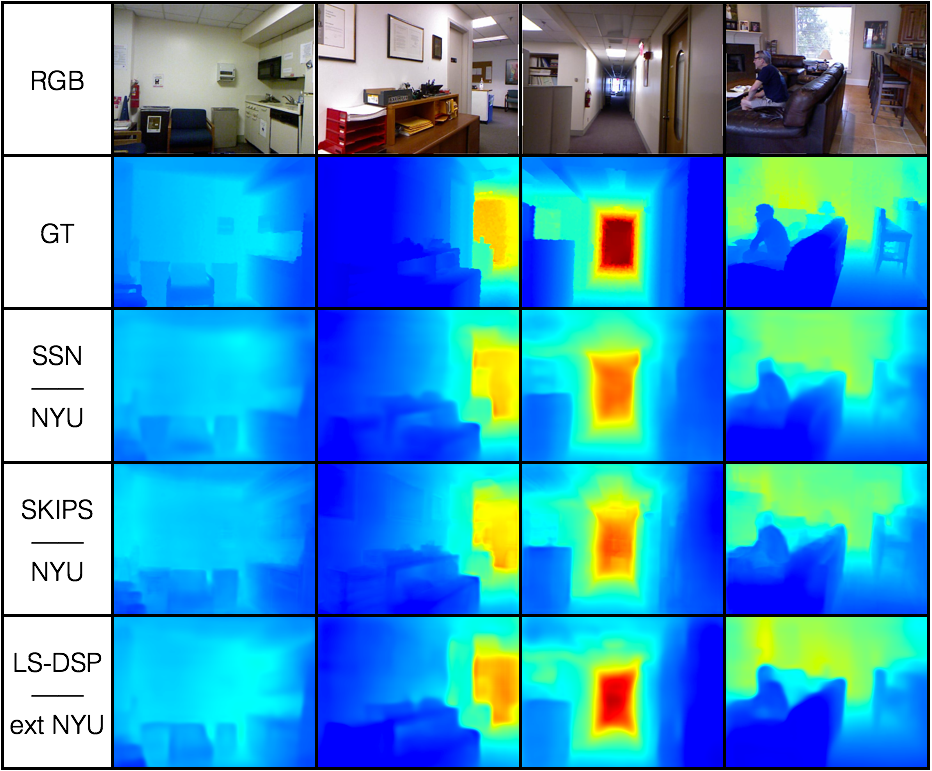}
\vspace{-1.5em}\caption{From top to bottom: RGB test image, depth ground-truth, SSN, Skips and LS-DSP outputs. Best viewed in color.}\vspace{-1em}
\label{qualitative}
\end{figure}

\vspace{1em}\noindent{\bf Qualitative evaluation.}
Fig.~\ref{qualitative} allows some qualitative comparisons of the single-scale network and two multi-scale ones (Skips and DSP). We observed that the methods using skip connections often produce sharper outputs, which is visible on the edge of the furniture (2nd column), or the ridges on the ceiling (1st column). Skip connections at the end of the network give sharpest inferences, at the price of artifacts (edges of the RGB images transferred to the depth map), visible on the paintings (2nd column, 4th row). We made similar observations for FPO whose architecture is  close to Skips, and, to lesser extent as the skips are in the middle layer of the network, with MSML. We also show our best results using the DSP trained on the extended NYU dataset. We see that the network generalizes well and produces better depth map than SSN. We can see, for instance, better inferences of farther distances (3rd columns) and better defined shapes, \eg the head of the person (4th column).

\begin{table}
\begin{center}
\resizebox{\columnwidth}{!}{%
\begin{tabular}{|c||c|c|c||c|c|c|}
\hline
& \multicolumn{3}{|c||}{\textbf{lower is better}} & \multicolumn{3}{|c|}{\textbf{higher is better}}\\
\hline\hline
Method & rel & log & rms & $\delta_1$ & $\delta_2$ & $\delta_3$ \\
\hline\hline

Eigen \cite{eigen2015predicting} & 0.158 & - & 0.641 & 76.9 & 95.0 & 98.8 \\

Li \cite{li2017two} & 0.143 & 0.063 & 0.635 & 78.8 & 95.8 & 99.1 \\

Laina \cite{laina2016deeper} & 0.138 & 0.060 & 0.592 & 78.5 & 95.2 & 98.7 \\
Cao \cite{cao2017estimating} & 0.141 & 0.060 & \textbf{0.540} & 81.9 & 96.5 & 99.2\\\hline\hline


LS-DSP & \textbf{0.133} & \textbf{0.057} & 0.569 & \textbf{83.0} & \textbf{96.6} & \textbf{99.3} \\

\hline
\end{tabular}%
}
\end{center}
\vspace{-1.5em}\caption{Comparison of our LS-DSP network trained on the extended dataset, against state-of-the-art results on NYU.}
\label{sota_methods}
\end{table}

\section{Conclusions}

This paper investigates the use of multi-scale information for depth inference, a question that has not been discussed so far in the literature. The presented experiments show that, on the official NYU dataset with 795 training images, the proposed multi-scale architectures consistently outperform the counterpart single scale architecture. When trained on a larger training set, of a size comparable to the latest published methods, the proposed DSP method gives impressive state-of-the art results while, in addition, the inferred depth maps look qualitatively better. 

\clearpage
\bibliographystyle{IEEEbib}
\bibliography{egbib}

\end{document}